\documentclass[runningheads]{llncs}
\usepackage{comment}
\usepackage{graphicx}
\usepackage{amsmath,amssymb} 
\usepackage{color}

\usepackage{float}
\usepackage{tabulary}
\usepackage{multirow}
\usepackage{booktabs}  
\usepackage{caption}
\usepackage{subfigure}
\usepackage{fancyhdr,graphicx}
\usepackage{diagbox}
\usepackage{bbding}
\usepackage{footnote}
\usepackage[lined,ruled,linesnumbered]{algorithm2e}
\usepackage[pagebackref=true,breaklinks=true,letterpaper=true,colorlinks,bookmarks=false]{hyperref}

\begin{document}
\pagestyle{headings}
\mainmatter
\def\ECCVSubNumber{3162}

\title{Towards Fast, Accurate and Stable \\ 3D Dense Face Alignment} 
\titlerunning{Towards Fast, Accurate and Stable 3D Dense Face Alignment}

\author{Jianzhu Guo\inst{1,2}$^{\rm \star}$\orcidID{0000-0002-8493-3689} \and
Xiangyu Zhu\inst{1,2}\thanks{Equal contribution.} \orcidID{0000-0002-4636-9677} \and\\
Yang Yang\inst{1,2}\orcidID{0000-0003-0559-5464} \and
Fan Yang\inst{3}\orcidID{0000-0003-4348-3148} \and\\
Zhen Lei\inst{1,2}\thanks{Corresponding author.}\orcidID{0000-0002-0791-189X} \and
Stan Z. Li\inst{4}\orcidID{0000-0002-2961-8096}}
\authorrunning{J. Guo, X. Zhu, Y. Yang, F. Yang, Z. Lei and S. Li}

\institute{
CBSR\&NLPR, Institute of Automation, Chinese Academy of Sciences \\ \and
School of Artificial Intelligence, University of Chinese Academy of Sciences\\ \and
College of Software, Beihang University\\ \and
School of Engineering, Westlake University\\
\email{\{jianzhu.guo,xiangyu.zhu,yang.yang,zlei,szli\}@nlpr.ia.ac.cn, fanyang@buaa.edu.cn}
}









\maketitle
\begin{abstract}
    Existing methods of 3D dense face alignment mainly concentrate on accuracy, thus limiting the scope of their practical applications.
    In this paper, we propose a novel regression framework named 3DDFA-V2 which makes a balance among speed, accuracy and stability.
    Firstly, on the basis of a lightweight backbone, we propose a meta-joint optimization strategy to dynamically regress a small set of 3DMM parameters, which greatly enhances speed and accuracy simultaneously.
	To further improve the stability on videos, we present a virtual synthesis method to transform one still image to a short-video which incorporates in-plane and out-of-plane face moving.
    On the premise of high accuracy and stability, 3DDFA-V2 runs at over 50fps on a single CPU core and outperforms other state-of-the-art heavy models simultaneously. Experiments on several challenging datasets validate the efficiency of our method.
    Pre-trained models and code are available at \url{https://github.com/cleardusk/3DDFA_V2}.
\end{abstract}

\section{Introduction}
3D dense face alignment is essential for many face related tasks, e.g., recognition~\cite{taigman2014deepface,booth20163d,guo2020learning,guo2018face,cao2020informative,xu2021searching}, animation~\cite{cao20133d}, avatar retargeting~\cite{cao2018stabilized}, tracking~\cite{xiong2015global}, attribute classification~\cite{bettadapura2012face,guo2017multi,guo2018dominant}, image restoration~\cite{yang2013structured,cao2019towards,cao2018learning}, anti-spoofing~\cite{wang2020deep,yu2020searching,qin2019learning,yu2020face,guo2019improving}. Recent studies are mainly divided into two categories: 3D Morphable Model (3DMM) parameters regression~\cite{jourabloo2016large,zhu2016face,liu2016joint,liu2017dense,tuan2017regressing,zhu2019face,3ddfa_cleardusk} and dense vertices regression~\cite{jackson2017large,feng2018joint}.
Dense vertices regression methods directly regress the coordinates of all the 3D points (usually more than 20,000) through a fully convolutional network~\cite{jackson2017large,feng2018joint}, achieving the state-of-the-art performance.
However, the resolution of reconstructed faces relies on the size of the feature map and these methods rely on heavy networks like hourglass~\cite{newell2016stacked} or its variants, which are slow and memory-consuming in inference.
The natural way of speeding it up is to prune channels. We try to prune 77.5\% channels on the state-of-the-art PRNet~\cite{feng2018joint} to achieve real-time speed on CPU, but find the error greatly increases 44.8\% (3.62\% vs. 5.24\%).
Besides, an obvious disadvantage is the presence of checkerboard artifacts due to the deconvolution operators, which is present in the supplementary material.
\begin{figure}[!h]
  \centering
  \includegraphics[width=0.8\textwidth]{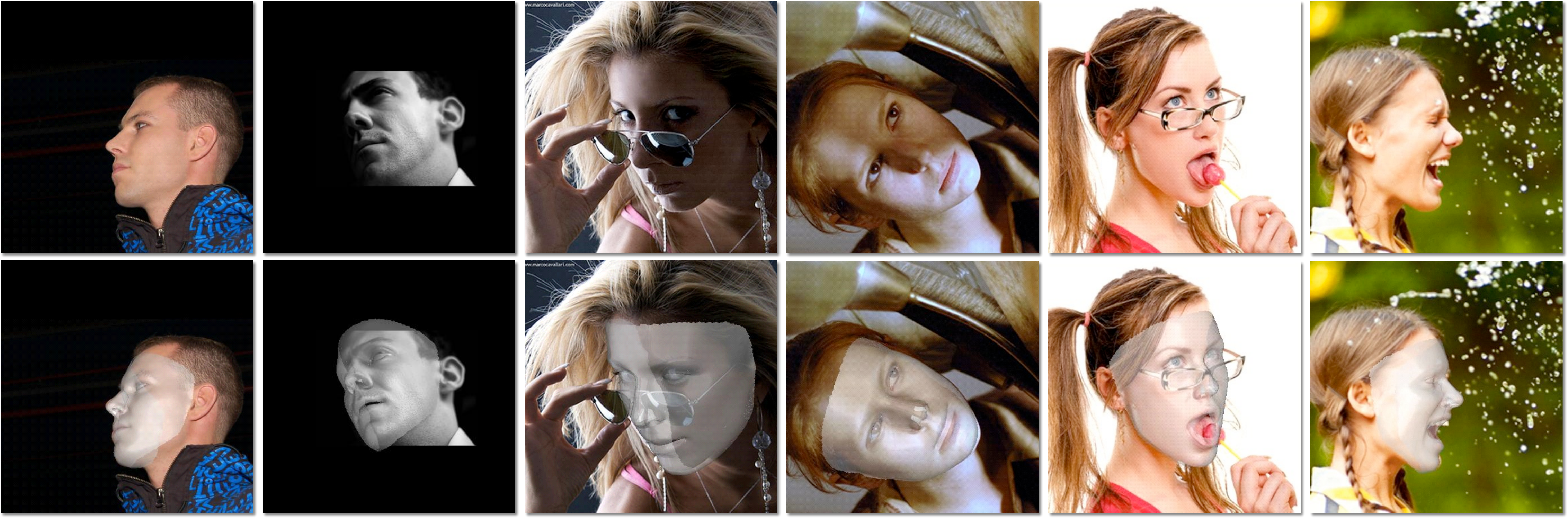}
  \caption{A few results from our \textit{3DDFA-V2 (M+R+S)} model, which runs at over 50fps on a single CPU core or over 130fps on multiple CPU cores.}
  \label{fig_demo_v4}
\end{figure}
Another strategy is to regress a small set of 3DMM parameters (usually less than 200). Compared with dense vertices, 3DMM parameters have low dimensionality and low redundancy, which are appropriate to regress by a lightweight network.
However, different 3DMM parameters influence the reconstructed 3D face~\cite{zhu2016face} differently, making the regression challenging since we have to dynamically re-weight each parameter according to their importance during training. Cascaded structures~\cite{zhu2016face,liu2017dense,zhu2019face} are always adopted to progressively update the parameters but the computation cost is increased linearly with the number of cascaded stages.

In this paper, we aim to accelerate the speed to CPU real time and achieve the state-of-the-art performance simultaneously. To this end, we choose to regress 3DMM parameters with a fast backbone, e.g. MobileNet. To handle the optimization problem of the parameters regression framework, we exploit two different loss terms WPDC and VDC~\cite{zhu2016face} (see Sec.~\ref{sec_meta_joint}) and propose our meta-joint optimization to combine the advantages of them.
The meta-joint optimization looks ahead by $k$-steps with WPDC and VDC on the meta-train batches, then dynamically selects the better one according to the error on the meta-test batch. By doing so, the whole optimization converges faster and achieves better performance than the vanilla-joint optimization.
Besides, a landmark-regression regularization is introduced to further alleviate the optimization problem to achieve higher accuracy.
In addition to single image, 3D face applications on videos are becoming more and more popular~\cite{cao20133d,cao2018stabilized,kim2018deep,kim2019neural}, where reconstructing stable results across consecutive frames is important, but it is often ignored by recent methods~\cite{zhu2016face,jackson2017large,feng2018joint,zhu2019face}.
Video-based training~\cite{peng2016recurrent,liu2018two,dong2018supervision,tai2018towards} is always adopted to improve the stability in 2D face alignment.
However, no video databases are publicly available for 3D dense face alignment.
To address it, we propose a 3D aided short-video-synthesis method, which simulates both in-plane and out-of-plane face moving to transform one still image to a short video, so that our network can adjust results of consecutive frames.
Experiments show our short-video-synthesis method significantly improves the stability on videos.

In general, our proposed framework 3DDFA-V2 are (i) \textit{fast}: It takes about 7.2ms with an single image as input (almost 24x faster than PRNet) and runs at over 50fps (19.2ms) on a single CPU core or over 130fps (7.2ms) on multiple CPU cores (i5-8259U processor), (ii) \textit{accurate}: By dynamically optimizing 3DMM parameters through a novel meta-optimization strategy combining the fast WPDC and VDC, we surpass the state-of-the-art results~\cite{zhu2016face,jackson2017large,feng2018joint,zhu2019face} under a strict computation burden in inference, and (iii) \textit{stable}: In a mini-batch, one still image is transformed slightly and smoothly into a short synthetic video, involving both in-plane and out-of-plane rotations, which provides temporal information of adjacent frames for training.
Extensive experimental results on four datasets show that the overall performance of our method is the best.

\section{Methodology}
\begin{figure}[!h]
  \centering
  \includegraphics[width=0.85\textwidth]{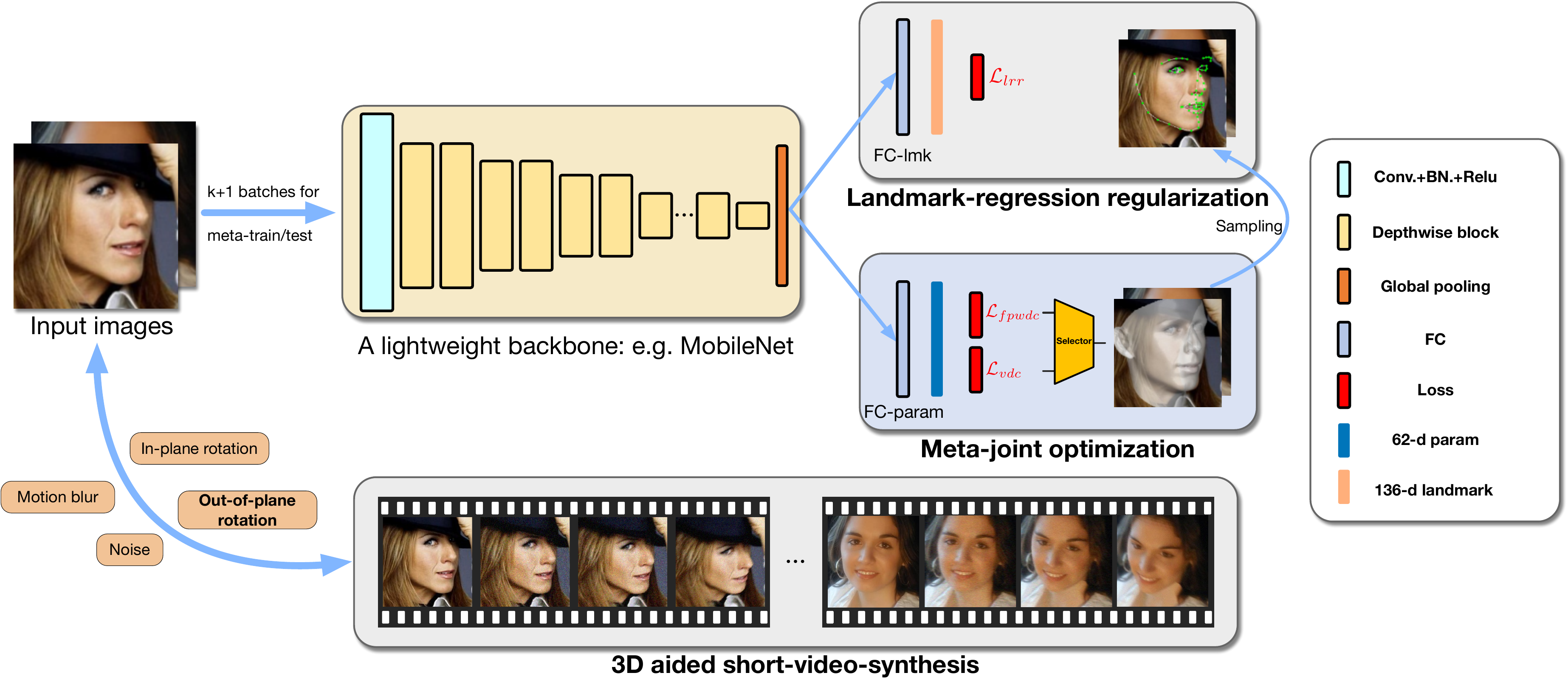}
  \caption{Overview of our 3DDFA-V2. Our architecture consists of four parts: the lightweight backbone like MobileNet for predicting 3DMM parameters, the meta-joint optimization of fWPDC and VDC, the landmark-regression regularization and the short-video-synthesis for training. The landmark-regression branch is discarded in inference, thus not increasing any computation burden.}
  \label{fig_guided}
\end{figure}

This section details our proposed approach. We first discuss 3D Morphable Model (3DMM)~\cite{blanz1999morphable}. Then, we introduce the proposed methods of the meta-joint optimization, landmark-regression regularization and 3D aided short-video-synthesis. The overall pipeline is illustrated in Fig.~\ref{fig_guided} and the algorithm is described in Algorithm~\ref{algo}.

\begin{algorithm}
    \scriptsize
        \caption{The overall algorithm of our proposed 3DDFA-V2.}
        \label{algo}
        \SetAlgoLined
        \SetKwInput{KwData}{Input}
        \SetKwInput{KwResult}{Init}
         \KwData{Training data $\mathcal{X} = \{ (x^l, p^l) \}_{l=1}^M$ .}
         \KwResult{Model parameters $\theta$ initialized randomly, the learning rate $\alpha$, look-ahead step $k$, length of 3D aided short-video-synthesis $n$, and batch-size of $B$.}
         \For{$i$ in max\_iterations}{
            Randomly sampling $k$ batches $\{ \mathcal{X}^{l}_{mtr} \}_{l=1}^k$ for meta-train and one disjoint batch $\mathcal{X}_{mte}$ for meta-test, each batch contains $B$ pairs: $\{ (x^l, p^l) \}_{l=1}^B$. \\
            \tcp{short-video-synthesis}
            \For{each $x \in \mathcal{X}_{mtr}$ \text{or} $\mathcal{X}_{mte}$}{
            	Synthesize a short-video with $n$ adjacent frames:  $\{(x_0, p_0|x_0) | x_0 = x\} \cup \ \{ (x'_j, p'_j|x'_j) |  x'_j = (M \circ P) (x_j) ,x_j =  (T \circ F)(x_{j-1}), 1 \leq j \leq n-1 \}$. \\
            }
            \tcp{Meta-joint optimization with landmark-regression regularization}

			Let $\theta^f_i, \theta^v_i \leftarrow \theta_{i}$; \\
            \For{$j = 1 \cdots k$}{
            	$\theta^f_{i+j} \leftarrow \alpha \nabla_{\theta^f_{i\text{+}j\text{-}1}} \left( \mathcal{L}_{fwpdc} (\theta^f_{i\text{+}j\text{-}1}, \mathcal{X}^j_{mtr}) + \frac{|l_{fwpdc}|} {|l_{lrr}|} \cdot \mathcal{L}_{lrr}(\theta^f_{i\text{+}j\text{-}1}, \mathcal{X}^j_{mtr}) \right)$;
            	$\theta^v_{i+j} \leftarrow \alpha \nabla_{\theta^v_{i\text{+}j\text{-}1}} \left(\mathcal{L}_{vdc} (\theta^v_{i\text{+}j\text{-}1}, \mathcal{X}^j_{mtr}) + \frac{|l_{vdc}|} {|l_{lrr}|} \cdot \mathcal{L}_{lrr}(\theta^v_{i\text{+}j\text{-}1}, \mathcal{X}^j_{mtr}) \right)$;
            }
            Select $\theta_{i\text{+}1} \leftarrow \arg\min_{\theta_{i+k}} \left( \mathcal{L}_{vdc}(\theta_{i+k}^f, \mathcal{X}_{mte}), \mathcal{L}_{vdc}(\theta_{i+k}^v, \mathcal{X}_{mte}) \right)$;

%
         }
    \end{algorithm}

\subsection{Preliminary of 3DMM}
\label{sec_sim_3dmm}
The original 3DMM can be described as:
\begin{equation}
	\mathbf{S} = \overline{\mathbf{S}} + \mathbf{A}_{id} \boldsymbol{\alpha}_{id} + \mathbf{A}_{exp} \boldsymbol{\alpha}_{exp},
\end{equation}
where $\mathbf{S}$ is the 3D face mesh, $\overline{\mathbf{S}}$ is the mean 3D shape, $\boldsymbol{\alpha}_{id}$ is the shape parameter corresponding to the 3D shape base $\mathbf{A}_{id}$, $\mathbf{A}_{exp}$ is the expression base and $\boldsymbol{\alpha}_{exp}$ is the expression parameter. After the 3D face is reconstructed, it can be projected onto the image plane with the scale orthographic projection:
\begin{equation}
\label{eq_scale_ortho_proj}
V_{2d}{(\mathbf{p})}=f * \mathbf{Pr} * \mathbf{R} *\left( \overline{\mathbf{S}}+\mathbf{A}_{id} \boldsymbol{\alpha}_{id}+\mathbf{A}_{exp} \boldsymbol{\alpha}_{exp} \right) + \mathbf{t}_{2d},
\end{equation}
where $V_{2d}{(\mathbf{p})}$ is the projection function generating the 2D positions of model vertices, $f$ is the scale factor, $\mathbf{Pr}$ is the orthographic projection matrix, $\mathbf{R}$ is the rotation matrix constructed by Euler angles including pitch, yaw, roll and $\mathbf{t}_{2d}$ is the translation vector. The complete parameters of 3DMM are $\mathbf{p} = [f, \mathrm{pitch}, \mathrm{yaw}, \mathrm{roll}, \mathbf{t}_{2d}, \boldsymbol{\alpha}_{id}, \boldsymbol{\alpha}_{exp}]$.

However, the three Euler angles will cause the gimbal lock~\cite{lepetit2005monocular} when faces are close to the profile view. This ambiguity will confuse the regressor to degrade the performance, so we choose to regress the similarity transformation matrix instead of $[f, \mathrm{pitch}, \mathrm{yaw}, \mathrm{roll}, \mathbf{t}_{2d}]$ to reduce the regression difficulty: $\mathbf{T} = f  \left[ \mathbf{R};  \mathbf{t}_{3d} \right]$,
where $\mathbf{T} \in \mathbb{R}^{3 \times 4}$ is constructed by a scale factor $f$, a rotation matrix $\mathbf{R}$ and a translation vector $\mathbf{t}_{3d} = \begin{bmatrix} \mathbf{t}_{2d} \\ 0 \end{bmatrix}$.
Therefore, the scale orthographic projection in Eqn.~\ref{eq_scale_ortho_proj} can be simplified as:
\begin{equation}
	V_{2d}({\mathbf{p}}) = \mathbf{Pr} * \mathbf{T} *  \begin{bmatrix} \overline{\mathbf{S}} + \mathbf{A} \boldsymbol{\alpha} \\ \mathbf{1} \end{bmatrix},
\end{equation}
where $\mathbf{A} = [\mathbf{A}_{id}, \mathbf{A}_{exp}]$ and $\boldsymbol{\alpha} = [\boldsymbol{\alpha}_{id}, \boldsymbol{\alpha}_{exp}]$. Our regression objective is described as $\mathbf{p} = [\mathbf{T}, \boldsymbol{\alpha}]$.

The high-dimensional parameters $\boldsymbol{\alpha}_{shp} \in \mathbb{R}^{199}$, $\boldsymbol{\alpha}_{exp} \in \mathbb{R}^ {29}$ are redundant, since 3DMM models the 3D face shape with PCA and the last parts of parameters have little effect on the face shape.
We choose only the first 40 dimensions of $\boldsymbol{\alpha}_{shp}$ and the first 10 dimensions of $\boldsymbol{\alpha}_{exp}$ as our regression target, since the NME increase is acceptable and the reconstruction can be greatly accelerated. The NME error heatmap caused by different size of shape and expression dimensions is present in the supplementary material. Therefore, our complete regression target is simplified as $\mathbf{p} = [\mathbf{T}^{3 \times 4}, \boldsymbol{\alpha}^{50}]$, with 62 dimensions in total, where $\boldsymbol{\alpha} = [\boldsymbol{\alpha}_{shp} ^{40}, \boldsymbol{\alpha}_{exp}^{10}]$.
To eliminate the negative impact of magnitude differences between $\mathbf{T}$ and $\boldsymbol{\alpha}$, Z-score normalizing is adopted: $\mathbf{p} = (\mathbf{p} - \boldsymbol{\mu}_{p}) / \boldsymbol{\sigma}_{p}$,
where $\boldsymbol{\mu}_{p} \in \mathbb{R} ^ {62}$ is the mean of parameters and $\boldsymbol{\sigma}_{p} \in \mathbb{R} ^ {62}$ indicates the standard deviation of parameters.

\subsection{Meta-joint Optimization} 
\label{sec_meta_joint}
We first review the Vertex Distance Cost (VDC) and Weighted Parameter Distance Cost (WPDC) in~\cite{zhu2016face}, then derivate the meta-joint optimization to facilitate the parameters regression.

The VDC term $\mathcal{L}_{vdc}$ directly optimizes $\mathbf{p}$ by minimizing the vertex distances between the fitted 3D face and the ground truth:
\begin{equation}
\label{eq_vdc}
	\mathcal{L}_{vdc} = \left\| V_{3d}\left(\mathbf{p} \right) - V_{3d}\left(\mathbf{p}^{g}\right) \right\|^{2},
\end{equation}
where $\mathbf{p}^g$ is the ground truth parameter, $\mathbf{p}$ is the predicted parameter and $V_{3d} (\cdot)$ is the 3D face reconstruction formulated as:
\begin{equation}
\label{eq_v3d}
	V_{3d}({\mathbf{p}}) = \mathbf{T} * \begin{bmatrix} \overline{\mathbf{S}} + \mathbf{A} \boldsymbol{\alpha} \\ \mathbf{1} \end{bmatrix}.
\end{equation}

Different from VDC, the WPDC term~\cite{zhu2016face} $\mathcal{L}_{wpdc}$ assigns different weights to each parameter:
\begin{equation}
\label{eq_wpdc}
	\mathcal{L}_{wpdc} = \left\| \mathbf{w} \cdot (\mathbf{p} - \mathbf{p}^{g}) \right\|^2,
\end{equation}
where $\mathbf{w}$ indicates the importance weight as follows:

\begin{equation}
\label{eq_wpdc_weight}
	\begin{aligned}
		\mathbf{w} &= \left( w_1, w_2, \dots, w_i, \dots, w_n \right), \\
		w_i &= \left\| V_{3d}(\mathbf{p}^{de,i}) - V_{3d}(\mathbf{p}^g)\right\| / Z, \\
		\mathbf{p}^{de,i} &= \left( \mathbf{p}_1^g, \mathbf{p}_2^g, \dots, \mathbf{p}_i, \dots, \mathbf{p}_n^g \right), \\
	\end{aligned}
\end{equation}
where $n$ is the number of parameters ($n = $ 62 in our regression framework), $\mathbf{p}^{de, i}$ is the $i$-degraded parameter whose $i$-th element is from the predicted $\mathbf{p}$, $Z$ is the maximum of $\mathbf{w}$ for regularization. The term $\left\| V_{3d}(\mathbf{p}^{de,i}) - V_{3d}(\mathbf{p}^g)\right\|$ models the importance of $i$-th parameter.

    
\begin{algorithm}[H]
    \scriptsize
    \caption{fWPDC: Fast WPDC Algorithm of 3DDFA-V2.}
    \label{fwpdc}
    \SetKwInOut{Input}{Input}\SetKwInOut{Output}{Output}
    \Input{ Shape and expression base: $\mathbf{A} = \left[ \mathbf{A}_{id}, \mathbf{A}_{exp} \right] \in \mathbb{R}^{3N \times 50}$ \\
        ~Mean shape: $\overline{\mathbf{S}} \in \mathbb{R}^{3 \times N}$ \\
        ~Predicted parameters: $\mathbf{p} = [\mathbf{T} \in \mathbb{R}^{3 \times 4}, \boldsymbol{\alpha} \in \mathbb{R}^{50}]$ \\
        ~Ground truth parameters: $\mathbf{p}^g = [\mathbf{T}^g \in \mathbb{R}^{3 \times 4}, \boldsymbol{\alpha}^g \in \mathbb{R}^{50}]$ \\
        ~Scale factor scalar: $f$ \\
        }
    \Output{ WPDC item} 
    \SetAlgoLined
    Initialize the weights of the parameter $\mathbf{T}$ and $\boldsymbol{\alpha}$: $\mathbf{w}_T \in \mathbb{R}^{3 \times 4}$, $\mathbf{w}_\alpha \in \mathbb{R}^{50}$\;
    \tcp{Calculating the weight of transform matrix}
    Reconstruct the vertices without projection: $\mathbf{S} = \overline{\mathbf{S}} + \mathbf{A} \boldsymbol{\alpha}^g \in \mathbb{R}^{3 \times N}$\;
    \For{$i = 1,2,3$}{
        $\mathbf{w}_T (:, i) = \big (\mathbf{T}(:,i)-\mathbf{T}^g(:,i) \big ) \cdot \left\| \mathbf{S} \left(i, : \right) \right\|$\;
    }
    $\mathbf{w}_T (:, 4) = \big (\mathbf{T}(:,4)-\mathbf{T}^g(:,4) \big ) \cdot \sqrt{N}$\ and then flatten $\mathbf{w}_T$ to the vector form in row-major order\;
    \tcp{Calculating the weight of shape and expression parameters}
    \For{$i = 1 \ldots 50$}{
        $\mathbf{w}_\alpha (i) = f \cdot \big( \boldsymbol{\alpha}(i) - \boldsymbol{\alpha}^g(i) \big) \cdot \left\| \mathbf{A}\left( :,i \right) \right\|$\;
    }
    \tcp{Calculating the fWPDC item}
    Get the maximum value $Z$ of the weights $\left( \mathbf{w}_T, \mathbf{w}_\alpha \right)$\ and normalize them: $\mathbf{w}_T = \mathbf{w}_T / Z$, $\mathbf{w}_\alpha = \mathbf{w}_\alpha / Z$\;
    Calculate the WPDC item: $\mathcal{L}_{fwpdc} = \left\|  \mathbf{w}_T \cdot \left( \mathbf{T} - \mathbf{T}^g \right) \right\|^2 + \left\| \mathbf{w}_\alpha \cdot  \left( \boldsymbol{\alpha} - \boldsymbol{\alpha}^g \right) \right\|^2$
    
\end{algorithm}

\textbf{fWPDC.} The original calculation of $\mathbf{w}$ in WPDC is rather slow as the calculation of each $w_i$ needs to reconstruct all the vertices once, which is a bottleneck for fast training.
We find that the vertices can be only reconstructed once by decomposing the weight calculation into two parts: the similarity transformation matrix $\mathbf{T}$, and the combination of shape and expression parameters $\boldsymbol{\alpha}$.
Therefore, we design a fast implementation of WPDC named fWPDC: (i) reconstructing the vertices without projection $\mathbf{S} = \overline{\mathbf{S}} + \mathbf{A} \boldsymbol{\alpha}$ and calculating $\mathbf{w}_T$ using the norm of row vectors; (ii) calculating $\mathbf{w}_{\alpha}$ using the norm of column vectors of $\mathbf{A}$ and the input scale $f$: $\mathbf{w}_\alpha (i) = f \cdot \big( \boldsymbol{\alpha}(i) - \boldsymbol{\alpha}^g(i) \big) \cdot \left\| \mathbf{A}\left( :,i \right) \right\|$;
(iii) Combining them to calculate the final cost.
The detailed algorithm of fWPDC is described in Algorithm~\ref{fwpdc}.
fWPDC only reconstructs dense vertices once, not 62 times as WPDC, thus greatly reducing the computation cost.
With 128 samples as a batch input, the original WPDC takes 41.7ms while fWPDC only takes 3.6ms. fWPDC is over 10x faster than the original WPDC while preserving the same outputs.

\begin{figure}[!htb]
  \centering
  \includegraphics[width=0.55\textwidth]{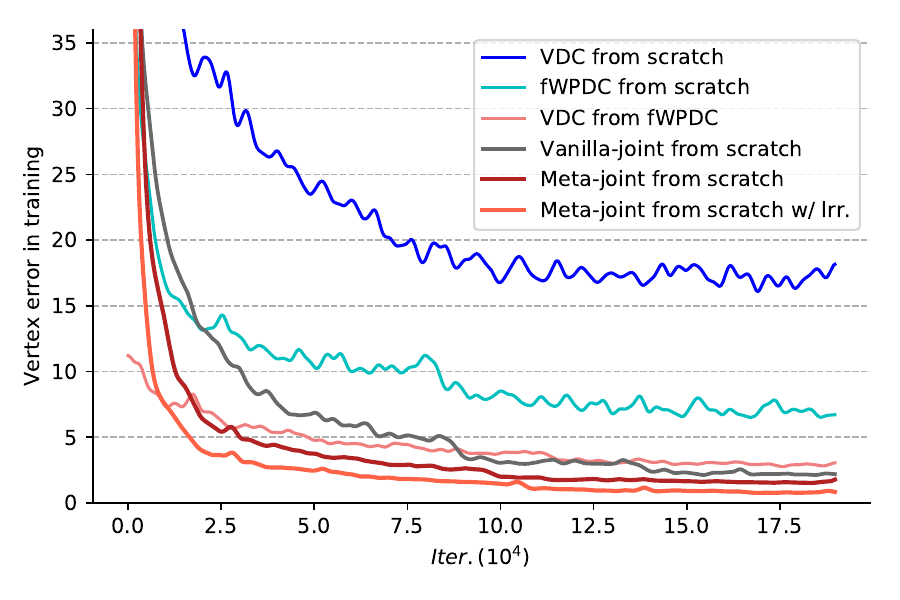}
  \caption{The vertex error in training on 300W-LP supervised by different loss terms. VDC from scratch has the highest error, fWPDC is lower than VDC, and VDC from fWPDC is better than both. When combining VDC and fWPDC, the proposed meta-joint optimization converges faster and reaches lower error than vanilla-joint, and achieves even better convergence when incorporating the landmark-regression regularization.
 }
  \label{fig_cascade}
\end{figure}

%

\textbf{Exploitation of VDC and fWPDC.} Through Eqn.~\ref{eq_vdc} and Eqn.~\ref{eq_wpdc}, we find: WPDC/fWPDC is suitable for parameters regression since each parameter is appropriately weighted, while VDC can directly reflect the goodness of the 3D face reconstructed from parameters.
In Fig.~\ref{fig_cascade}, we investigate how these two losses converge as the training progresses.
It is shown that the optimization is difficult for VDC since the vertex error is still over 15 when training converges. The work in~\cite{zhu2019face} also demonstrates that optimizing VDC with gradient descent converges very slowly due to the "zig-zagging" problem.
In contrast, the convergence of fWPDC is much faster than VDC and the error is about 7 when training converges.
Surprisingly, if the fWPDC-trained model is fine-tuned by VDC, we can get a much lower error than fWPDC.
Based on the above observation, we conclude that: \textit{training from scratch with VDC is hard to converge} and \textit{the network is not fully trained by fWPDC in the late stage}.

\begin{figure}[!h]
  \centering
  \includegraphics[width=0.9\textwidth]{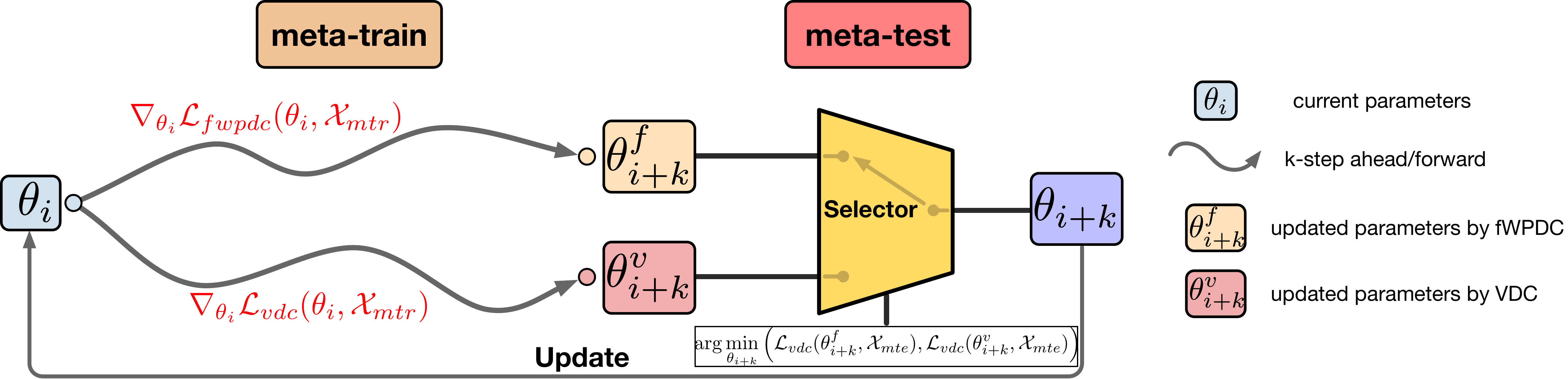}
  \caption{Overview of the meta-joint optimization.}
  \label{fig_meta_joint}
\end{figure}

\textbf{Meta-joint optimization.} Based on above discussions, it is natural to weight two terms to perform a vanilla-joint optimization: $\mathcal{L}_{vanilla\textit{-}joint}	= \beta \mathcal{L}_{fwpdc}\\ + (1 - \beta) \frac{|l_{fwpdc}|}{|l_{vdc}|} \cdot \mathcal{L}_{vdc}$,
where $\beta \in [0, 1]$ controls the importance between fWPDC and VDC. 
However, the vanilla-joint optimization relies on the manually set hyper-parameter $\beta$ and does not achieves satisfactory results in Fig.~\ref{fig_cascade}.
Inspired by Lookahead~\cite{zhang2019lookahead} and MAML~\cite{finn2017model}, we propose a meta-joint optimization strategy to dynamically combine fWPDC and VDC. The overview of the meta-joint optimization is shown in Fig.~\ref{fig_meta_joint}.
In the training process, the model looks ahead by $k$-steps with the cost fWPDC or VDC on $k$ meta-train batches $\mathcal{X}_{mtr}$, then selects the better one between fWPDC and VDC according to the vertex error on the meta-test batch.
Specifically, the whole meta-joint optimization consists of four steps: (i) sampling $k$ batches of training samples $\mathcal{X}_{mtr}$ for meta-train and one batch $\mathcal{X}_{mte}$ for meta-test; (ii) meta-train: updating the current model parameters $\theta_i$ with fWPDC and VDC on $\mathcal{X}_{mtr}$ by $k$-steps, respectively, getting two parameter states $\theta_{i+k}^f$ and $\theta_{i+k}^v$; (iii) meta-test: evaluating the vertex error $\theta_{i+k}^f$ and $\theta_{i+k}^v$ on $\mathcal{X}_{mte}$; (iv) selecting the parameters which have the lower error to update $\theta_i$.
The proposed meta-joint optimization can be directly embedded into the standard training regime.
From Fig.~\ref{fig_cascade}, we can observe that the meta-joint optimization converges faster than vanilla-joint and has the lower error.

\subsection{Landmark-regression Regularization}
\label{sec_guided}
In 3D face reconstruction~\cite{deng2019accurate,chinaev2018mobileface,tewari2017mofa,tewari2019fml,gecer2019ganfit}, the 2D sparse landmarks after projecting are usually used as an extra regularization to facilitate the parameters regression.
In our regression framework, we find that treating 2D sparse landmarks as a auxiliary regression task benefits more.
As shown in Fig.~\ref{fig_guided}, we add an additional landmark-regression task on the global pooling layer, trained by L2 loss.
The difference between the former landmark-regularization and the latter landmark-regression regularization is that the latter introduces extra parameters to regress the landmarks. In other words, the landmark-regression regularization is a task-level regularization.
From the tomato curve in Fig.~\ref{fig_cascade}, we get a lower error by incorporating the landmark-regression regularization.
The comparative results in Table~\ref{tab_abla} show our proposed landmark-regression regularization is better than landmark-regularization (3.59\% vs. 3.71\% on AFLW2000-3D).
The landmark-regression regularization is formulated as: $\mathcal{L}_{lrr} = \frac{1}{N} \sum_{i=1}^{N} \left\| l_i - l_i^g \right\| _2 ^2$, where $N$ is 136 here as we utilize 68 2D landmarks and flatten them into a 136-d vector.

\subsection{3D Aided Short-video-synthesis}
\label{sec_syn}
Video based 3D face applications have become more and more popular~\cite{cao20133d,cao2018stabilized,kim2018deep,kim2019neural} recently. In these applications, 3D dense face alignment methods are required to run on videos and provide stable reconstruction results across adjacent frames. The stability means that the changing of the reconstructed 3D faces across adjacent frames should be consistent with the true face moving in a fine-grained level.
However, most of existing methods~\cite{zhu2016face,zhu2019face,jackson2017large,feng2018joint} omit this requirenment and the predictions suffer from random jittering.
In 2D face alignment, post-processing like temporal filtering is a common strategy to reduce the jittering, but it degrades the precision and causes the frame delay.
Besides, since no public video databases for 3D dense face alignment are available,
the video training strategies~\cite{dong2018supervision,peng2016recurrent,tai2018towards,liu2018two} cannot work here. A challenge arises: \textit{can we improve the stability on videos with only still images available when training?}

\begin{figure}[!h]
  \centering
  \includegraphics[width=0.7\textwidth]{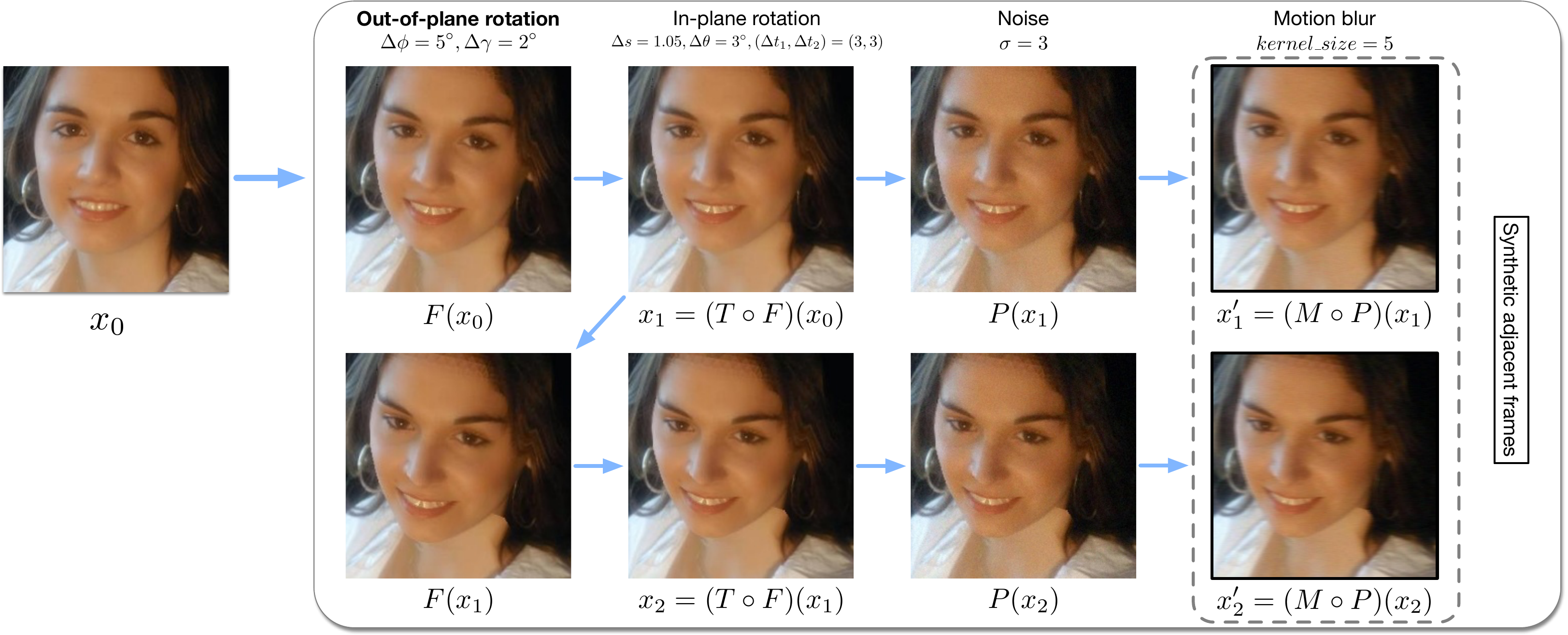}
  \caption{An illustration of how two adjacent frames are synthesized in our 3D aided short-video-synthesis.}
  \label{fig_3d_synthesis}
\end{figure}

To address this challenge, we propose a batch-level 3D aided short-video-synthesis strategy, which expands one still image to several adjacent frames, forming a short synthetic video in a mini-batch.
The common patterns in a video can be modelled as: (i) Noise. We model noise as $P(x) = x + \mathcal{N}(0, \Sigma)$, where $\Sigma = \sigma^2 I$. (ii) Motion Blur. Motion blur can be formulated as $M(x) = K \ast x$, where $K$ is the convolution kernel (the operator $\ast$ denotes a convolution). (iii) In-plane rotation. Given two adjacent frames $x_{t}$ and $x_{t+1}$, the in-plane temporal change from $x_t$ to $x_{t+1}$ can be described as a similarity transform $T ( \cdot )$:
\begin{equation}
    T(\cdot) = \Delta s \begin{bmatrix} \cos(\Delta \theta) & -\sin(\Delta \theta) & \Delta t_1  \\ \sin(\Delta \theta) & \cos(\Delta \theta) & \Delta t_2 \end{bmatrix},
\end{equation}
where $\Delta s$ is the scale perturbation, $\Delta \theta$ is the rotation perturbation, $\Delta t_1$ and $\Delta t_2$ are translation perturbations.
(iv) Since human faces share similar 3D structure, we are also able to synthesize the out-of-plane face moving. Face profiling~\cite{zhu2016face} $F (\cdot)$, which is originally proposed to solving large-pose face alignment, is utilized to progressively increase the yaw angle $\Delta \phi$ and pitch angle $\Delta \gamma$ of the face.
Specifically, we sample several still images in a mini-batch and for each still image $x_0$, we transform it slightly and smoothly to generate a synthetic video with $n$ adjacent frames: $\{ x'_j | x'_{j} = (M \circ P) (x_j), x_{j} = (T \circ F) (x_{j-1}), 1 \leq j \leq n-1 \} \cup \{x_0\}$. In Fig.~\ref{fig_3d_synthesis}, we give an illustration of how these transformations are applied on an image to generate several adjacent frames.

\begin{figure}[!htb]
  \centering
  \includegraphics[width=0.7\textwidth]{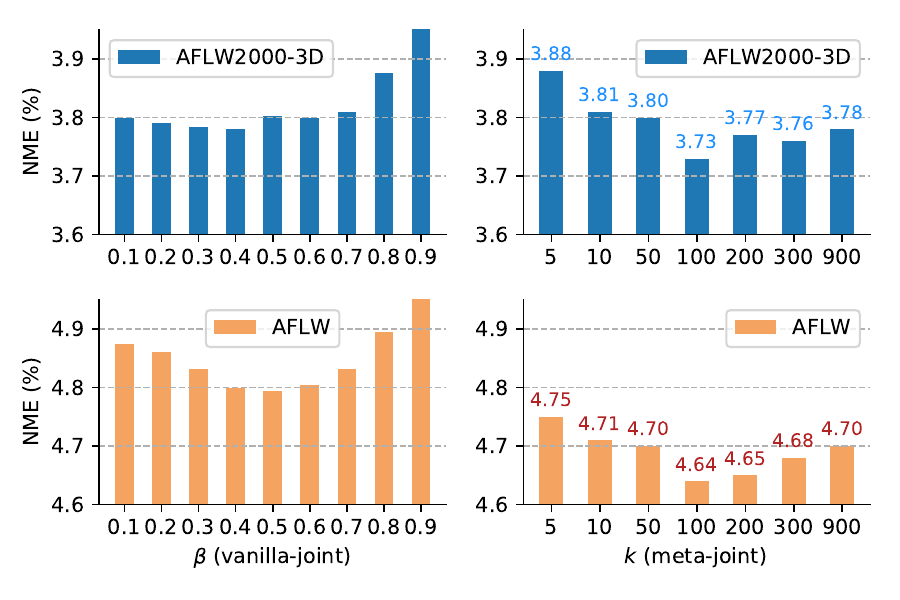}
  \caption{Ablative results of the vanilla-joint optimization with different $\beta$ and meta-joint optimization with different $k$. Lower NME (\%) is better.}
  \label{fig_abla}
\end{figure}

\begin{table}[!h]
\small
\centering
\caption{The NME (\%) of different methods on AFLW2000-3D and AFLW. The first and the second best results are highlighted. M, R, S denote the meta-joint optimization, landmark-regression regularization and short-video-synthesis, respectively.}
\resizebox{0.825\textwidth}{!} {
\begin{tabular}{c||c|c|c|c||c|c|c|c}
\hline
\multirow{2}{*}{\textbf{Method}} & \multicolumn{4}{c||}{\textbf{AFLW2000-3D (68 pts)}} & \multicolumn{4}{c}{\textbf{AFLW (21 pts)}} \\ \cline{2-9} 
 & {[}0, 30{]} & {[}30, 60{]} & {[}60, 90{]} & Mean & {[}0, 30{]} & {[}30, 60{]} & {[}60, 90{]} & Mean \\ \hline
 ESR~\cite{cao2014face} & 4.60 & 6.70 & 12.67 & 7.99 & 5.66 & 7.12 & 11.94 & 8.24 \\ 
SDM~\cite{xiong2015global} & 3.67 & 4.94 & 9.67 & 6.12 & 4.75 & 5.55 & 9.34 & 6.55 \\ 
3DDFA~\cite{zhu2016face} & 3.78 & 4.54 & 7.93 & 5.42 & 5.00 & 5.06 & 6.74 & 5.60 \\ 
3DDFA+SDM~\cite{zhu2016face} & 3.43 & 4.24 & 7.17 & 4.94 & 4.75 & 4.83 & 6.38 & 5.32 \\ 
Yu et al.~\cite{yu2017learning} & 3.62 & 6.06 & 9.56 & 6.41 & - & - & - & - \\ 
DeFA~\cite{liu2017dense} & - & - & - & 4.50 & - & - & - & - \\ 
3DSTN~\cite{bhagavatula2017faster} & 3.15 & 4.33 & 5.98 & 4.49 & \textbf{3.55} & \textbf{3.92} & 5.21 & \textbf{4.23} \\ 
3D-FAN~\cite{bulat2017far} & 3.15 & 3.53 & 4.60 & 3.76 & 4.40 & 4.52 & 5.17 & 4.69 \\
3DDFA-TPAMI~\cite{zhu2019face} & 2.84 & 3.57 & 4.96 & 3.79 & 4.11 & 4.38 & 5.16 & 4.55 \\ 
PRNet~\cite{feng2018joint} & \textbf{2.75} & 3.51 & 4.61 & 3.62 & 4.19 & 4.69 & 5.45 & 4.77 \\ \hline 
3DDFA-V2 (M+R) & \textbf{2.75} & \textbf{3.49} & \textbf{4.53} & \textbf{3.59} & 4.06 & 4.41 & \textbf{5.02} & 4.50 \\ 
3DDFA-V2 (M+R+S) & \textbf{2.63} & \textbf{3.42} & \textbf{4.48} & \textbf{3.51} & \textbf{3.98} & \textbf{4.31} & \textbf{4.99} & \textbf{4.43} \\ \hline
\end{tabular}
}
\label{tab_face_alignment}
\end{table}

\begin{table}[!h]
\small
\centering
\caption{The NME (\%) on Florence, AFLW2000-3D (Dense), NME (\%) / Stability (\%) on Menpo-3D, running complexity and time with different methods. Our method outputs 3D dense vertices with only 2.1ms (2ms for parameters prediction and 0.1ms for vertices reconstruction) in GPU or 7.2ms in CPU (6.2ms for parameters prediction and 1ms for vertices reconstruction). The first and second best results are highlighted.}
\resizebox{0.9\textwidth}{!} {
\begin{tabular}{c||c|c|c||c|c|c}
\hline
\textbf{Methods} & \textbf{Florence} & \textbf{\begin{tabular}[c]{@{}c@{}}AFLW2000-3D\\ (Dense)\end{tabular}} & \textbf{Menpo-3D} & \textbf{Params} & \textbf{MACs} & \textbf{Run Time (ms)} \\ \hline
3DDFA~\cite{zhu2016face} & 6.38 & 6.56 & - & - & - & 75.7=23.2(GPU)+52.5(CPU) \\ 
VRN~\cite{jackson2017large} & 5.27 & - & - & - & - & 69.0(GPU) \\ 
DeFA~\cite{liu2017dense} & - & 6.04 & - & - & 1426M & 35.4=11.8(GPU)+23.6(CPU) \\ 
PRNet~\cite{feng2018joint} & 3.76 & 4.41 & 1.90 / 0.54 & 13.4M & 6190M & 9.8(GPU) / 175.0(CPU) \\ \hline
3DDFA-V2 (M+R) & \textbf{3.59} & \textbf{4.20} & \textbf{1.86} / \textbf{0.52} & \multirow{2}{*}{\textbf{3.27M}} & \multirow{2}{*}{\textbf{183M}} & \multirow{2}{*}{\textbf{2.1(GPU)} / \textbf{7.2(CPU)}} \\ 
3DDFA-V2 (M+R+S) & \textbf{3.56} & \textbf{4.18} & \textbf{1.71} / \textbf{0.48} &  &  &  \\ \hline
\end{tabular}
}
\label{tab_dense_speed}
\end{table}

\section{Experiments}
In this section, we first introduce the datasets and protocols; then, we give comparison experiments on the accuracy and stability; thirdly, the complexity and running speed are evaluated; extensive discussions are finally made. The implementation details, generalization and scaling-up ability of our proposed method are in the supplementary material.

\subsection{Datasets and Evaluation Protocols}
Five datasets are used in our experiments:
\textbf{300W-LP}~\cite{zhu2016face} (300W Across Large Poses) is composed of the synthesized large-pose face images from 300W~\cite{sagonas2013300}, including AFW~\cite{zhu2012face}, LFPW~\cite{belhumeur2013localizing}, HELEN~\cite{zhou2013extensive}, IBUG~\cite{sagonas2013300}, and XM2VTS~\cite{messer1999xm2vtsdb}. Specifically, the face profiling method~\cite{zhu2016face} is adopted to generate 122,450 samples across large poses.
\textbf{AFLW}~\cite{koestinger2011annotated} consists of 21,080 in-the-wild faces (following ~\cite{zhu2019face,3ddfa_cleardusk}) with large poses (yaw from -90$^\circ$ to 90$^\circ$). Each image is annotated up to 21 visible landmarks.
\textbf{AFLW2000-3D}~\cite{zhu2016face} is constructed by~\cite{zhu2016face} for evaluating 3D face alignment performance, which contains the ground truth 3D faces and the corresponding 68 landmarks of the first 2,000 AFLW samples.
\textbf{Florence}~\cite{bagdanov2011florence} is a 3D face dataset containing 53 subjects with its ground truth 3D mesh acquired from a structured-light scanning system. For evaluation, we generate renderings with different poses for each subject following VRN~\cite{jackson2017large} and PRNet~\cite{feng2018joint}.
\textbf{Menpo-3D}~\cite{zafeiriou20173d} provides a benchmark for evaluating 3D facial landmark localization algorithms in the wild in arbitrary poses. Specifically, Menpo-3D provides 3D facial landmarks for 55 videos from 300-VW~\cite{shen2015first} competition.

\textbf{Protocols.} The protocol on AFLW follows~\cite{zhu2016face} and Normalized Mean Error (NME) by bounding box size is reported. Two protocols on AFLW2000-3D are applied: the first one follows AFLW, and the other one follows~\cite{feng2018joint} to evaluate the NME of 3D face reconstruction normalized by the bounding box size. For Florence, we follow~\cite{jackson2017large,feng2018joint} to evaluate the NME of 3D face reconstruction normalized by outer interocular distance. As for Menpo-3D, we evaluate the NME on still frames and the stability across adjacent frames. We calculate the stability following~\cite{tai2018towards} by measuring the NME between the predicted offsets and the ground-truth offsets of adjacent frames.
Specifically, at frame $t-1$ and $t$, the ground-truth landmark offset is $\Delta p =  p_t - p_{t-1}$, the prediction offset is $\Delta q = q_t - q_{t-1}$, the error $\Delta p - \Delta q$ normalized by the bounding box size represents the stability.
Since 300W-LP only has the indices of 68 landmarks, we use 68 landmarks of Menpo-3D for consistency.

\subsection{Ablation Study}

\begin{table}[!h]
    \centering
    \caption{The comparative and ablative results on AFLW2000-3D and AFLW. The mean NMEs (\%) across small, medium and large poses on AFLW2000-3D and AFLW are reported. lmk. indicates landmark constraint on the parameter regression like~\cite{tewari2017mofa} and lrr. is the proposed landmark-regression regularization.}
    \resizebox{0.7\textwidth}{!} {
    \begin{tabular}{c|c|c|c} \hline
     & \textbf{Method} & \textbf{AFLW2000-3D} & \textbf{AFLW} \\ \hline
    \multirow{2}{*}{Baseline} 
     & VDC & 5.23 & 6.37 \\ 
     & fWDPC & 4.04 & 5.10 \\ \hline
    \multirow{3}{*}{Joint-optimization Options} 
     & VDC from fWPDC & 3.88 & 4.83 \\
     & Vanilla-joint & 3.80 & 4.80 \\
     & \textbf{Meta-joint} & \textbf{3.73} & \textbf{4.64} \\ \hline 
    \multirow{4}{*}{Utilization of 2D landmarks} 
     & VDC w/ lrr. & 3.92 & 4.92 \\
     & fWPDC w/ lrr. & 3.89 & 4.84 \\
     & Meta-joint w/ lmk. & 3.71 & 4.80 \\ 
     & \textbf{Meta-joint w/ lrr.} & \textbf{3.59} & \textbf{4.50} \\ \hline
    \end{tabular}
    }
    \label{tab_abla}
\end{table}

\begin{table}[!h]
    \centering
    \caption{Comparisons of NME (\%) / Stability (\%) on Menpo-3D. svs. indicates short-video-synthesis, rnd. indicates applying in-plane and out-of-plane rotations randomly in one mini-batch.}
    \resizebox{0.4\textwidth}{!} {
    \begin{tabular}{c|c}
    \hline
    \textbf{Method} & \textbf{Menpo-3D} \\ \hline
    fWPDC w/o svs. & 1.96 / 0.54 \\
    fWPDC w/ svs. & \textbf{1.84 / 0.51} \\ \hline
    Meta-joint+lrr. w/o svs. & 1.86 / 0.52 \\
    Meta-joint+lrr. w/ rnd. & 1.76 / 0.50 \\
    Meta-joint+lrr. w/ svs.& \textbf{1.71} / \textbf{0.48} \\ \hline
    \end{tabular}
    }
    \label{tab_syn}
\end{table}

To evaluate the effectiveness of the meta-joint optimization and the landmark-regression regularization, we carry out comparative experiments including our two baselines: \textit{VDC} and \textit{fWPDC}, three joint options: (i) \textit{VDC from fWPDC}: fine-tune the model with VDC loss from the pre-trained model by fWPDC; (ii) \textit{Vanilla-joint}: weight VDC and fWPDC by the best scalar $\beta=0.5$; (iii) \textit{Meta-joint}: the proposed meta-joint optimization with best $k=100$ and four options of how the 2D landmarks are utilized.
From Table~\ref{tab_abla}, Table~\ref{tab_syn}, Fig.~\ref{fig_cascade} and Fig.~\ref{fig_abla}, we can draw the following conclusions:

\textbf{Meta-joint optimization performs better.}
Comparing with two baselines \textit{VDC} and \textit{fWPDC}, all three joint optimization methods perform better. Among three joint optimization methods, the proposed meta-joint performs better than \textit{VDC from fWPDC} and \textit{vanilla-joint}: the mean NME drops from $4.04\%$ to $3.73\%$ on AFLW2000-3D and $5.10\%$ to $4.64\%$ on AFLW when compared with the baseline \textit{fWPDC}.
Furthermore, we conduct ablative experiments with different $\beta$ for \textit{vanilla-joint} and different look-ahead step $k$ in Fig.~\ref{fig_abla}. We can observe that $\beta=0.5$ is the best setting for \textit{vanilla-joint}, but \textit{meta-joint} still outperforms it and $k=100$ performs best on both AFLW2000-3D and AFLW. Overall, the proposed meta-joint optimization is effective in alleviate the training and promoting the performance.

\textbf{Landmark-regression regularization benefits.}
Another contribution is the landmark-regression regularization, which can also be regarded as an auxiliary task to parameters regression.
From Table~\ref{tab_abla}, the improvements from \textit{fWPDC} to \textit{fWPDC w/ lrr.} on AFLW2000-3D and AFLW are $0.15\%$ and $0.26\%$, and the improvements from \textit{Meta-joint} to \textit{Meta-joint w/ lrr.} on AFLW2000-3D and AFLW are $0.14\%$ and $0.14\%$.
We also compare the proposed landmark-regression regularization with prior methods~\cite{tewari2017mofa,tewari2019fml} which directly impose landmark constraint on the parameter regression, the results show ours is significantly better: 3.59\% vs. 3.71\% on AFLW2000-3D.
We further evaluate the performance of the landmark-regression branch on AFLW2000-3D and AFLW. The performances are 3.58\% and 4.52\% respectively, which are close to the parameter branch. It indicates that these two tasks are highly related.
Overall, the landmark-regression regularization benefits the training and promotes the performance.


\textbf{Short-video-synthesis improves stability.}
The last contribution is 3D aided short-video-synthesis, which is designed to enhance stability on videos by augmenting one still image to a short video in a mini-batch.
The results in Table~\ref{tab_syn} indicate that short-video-synthesis works for both the fWPDC and meta-joint optimization.
With short-video-synthesis and landmark-regression regularization, the performance on still frames improves from $1.86\%$ to $1.71\%$ and the stability improves from $0.52\%$ to $0.48\%$.
We also evaluate the performance by randomly applying in-plane and out-of-plane rotations in each mini-batch and find it is worse than short-video-synthesis: 1.76\% / 0.50\% v.s. 1.71\% / 0.48\%.
These results validate the effectiveness of the 3D aided short-video-synthesis.

\subsection{Evaluations of Accuracy and Stability}
 
\textbf{Sparse Face Alignment.} We use AFLW2000-3D and AFLW to evaluate sparse face alignment performance with small, medium and large yaw angles. The results in Table~\ref{tab_face_alignment} indicate that our 3DDFA-V2 performs better than PRNet (3.51\% vs. 3.62\%) in AFLW2000-3D and better than 3DDFA-TPAMI~\cite{zhu2019face} in AFLW (4.43\% vs. 4.55\%).
Note that these results are achieved with only 3.27M parameters (24\% of PRNet) and it takes 6.2ms (3.5\% of PRNet) in CPU. The sampling of 68 / 21 landmarks from 3DMM is extremely fast, only 0.01ms (CPU), which can be ignored.


\textbf{Dense Face Alignment.} Dense face alignment is evaluated on Florence and AFLW2000-3D. Our evaluation settings follow~\cite{feng2018joint} to keep consistency. The results in Table~\ref{tab_dense_speed} show that our 3DDFA-V2 significantly outperforms others. As for 3D dense vertices reconstruction, 45K vertices only takes 1ms in CPU (0.1ms in GPU) with our regression framework.



\textbf{Video-based 3D Face Alignment.} We use Menpo-3D to evaluate both the accuracy and stability. Table~\ref{tab_syn} has already shown the superiority of short-video-synthesis. We choose to compare our method with recent PRNet~\cite{feng2018joint} in Table~\ref{tab_dense_speed}. The results indicate that our method significantly surpasses PRNet in both the accuracy and stability on videos of Menpo-3D with a much lower computation cost.



\subsection{Evaluations of Speed}
We compare parameter numbers, MACs (Multiply-Accumulates) measuring the number of fused Multiplication and Addition operations, and the running time of our method with others in Table~\ref{tab_dense_speed}.
As for the running speed, 3DDFA~\cite{zhu2016face} takes 23.2ms (GPU) for predicting parameters and 52.5ms (CPU) for PNCC construction, DeFA~\cite{feng2018joint} needs 11.8ms (GPU) to predict 3DMM parameters and 23.6ms (CPU) for post-processing, VRN~\cite{jackson2017large} detects 68 2D landmarks with 28.4ms (GPU) and regresses the 3D dense vertices with 40.6ms (GPU), PRNet~\cite{feng2018joint} predicts the 3D dense vertices with 9.8ms (GPU) or 175ms (CPU). Compared with them, our 3DDFA-V2 takes only 2ms (GPU) or 6.2ms (CPU) to predict 3DMM parameters and 0.1ms (GPU) or 1ms (CPU) to reconstruct 3D dense vertices.

Specifically, compared with the recent PRNet~\cite{feng2018joint}, the parameters of our 3DDFA-V2 (3.27M) are less than one-quarter of PRNet (13.4M), and the MACs are less than $1 / 30$ (183.5M vs. 6190M).
We measure the overall running time on GeForce GTX 1080 GPU and i5-8259U CPU with 4 cores. Note that our 3DDFA-V2 takes only 7.2ms, which is almost 24x faster than PRNet (175ms).
Besides, we benchmark our 3DDFA-V2 on a single CPU core (using only one thread) and \textit{the running speed of our method is about 19.2ms (over 50fps), including the reconstruction time}. The specific CPU configuration is i5-8259U CPU @ 2.30GHz on a 13-inch MacBook Pro.

\subsection{Analysis of Meta-joint Optimization}
\begin{figure}[!htb]
  \centering
  \includegraphics[width=0.9\textwidth]{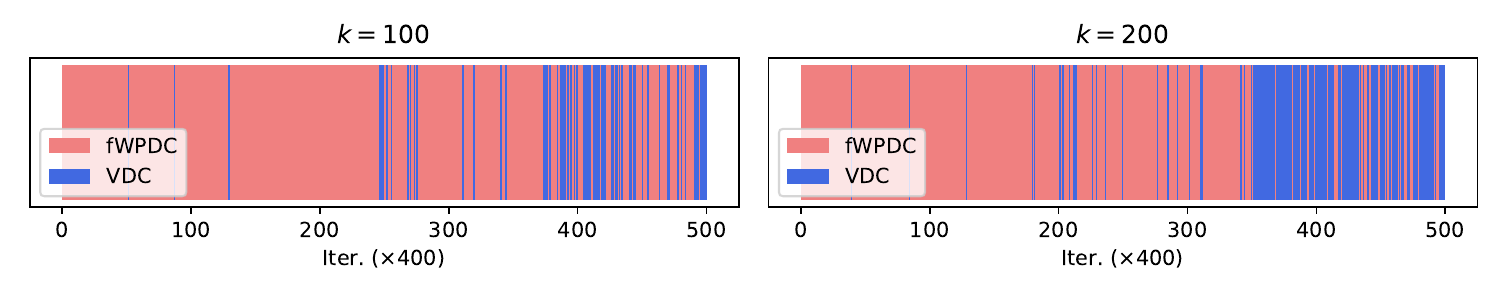}
  \caption{Auto-selection result of the selector in the meta-joint optimization.}
  \label{fig_metalk}
\end{figure}

We visualize the auto-selection result of fWPDC and VDC in the meta-joint optimization, as shown in Fig.~\ref{fig_metalk}. We can observe that both $k=100$ and $k=200$ show the same trend: fPWDC dominates in the early stage and VDC guides in the late stage. This trend is consistent with the previous observations and gives a clear description of why our proposed meta-joint optimization works.

\section{Conclusion}
In this paper, our proposed 3DDFA-V2 has successfully pursued the fast, accurate and stable 3D dense face alignment simultaneously. Towards this target, we make three main efforts: (i) proposing a fast WPDC named fWPDC and the meta-joint optimization to combine fWPDC and VDC to alleviate the problem of optimization; (ii) imposing an extra landmark-regression regularization to promote the performance to state-of-the-art; (iii) proposing the 3D aided short-video-synthesis method to improve the stability on videos.
The experimental results demonstrate the effectiveness and efficiency of our proposed methods. Our promising results pave the way for real-time 3D dense face alignment in practical use and the proposed methods may improve the environment by reducing the amount of carbon dioxide released by the huge amounts of energy consumed by GPUs.

\section{Acknowledgement}
This work was supported in part by the National Key Research \& Development Program (No. 2020YFC2003901), Chinese National Natural Science Foundation Projects \#61872367, \#61876178, \#61806196, \#61976229.


\clearpage

\bibliographystyle{splncs04}
\bibliography{egbib}

\clearpage
\appendix
\section*{A. Checkerboard Artifacts}
\begin{figure}[!htb]
  \centering
  \includegraphics[width=0.7\textwidth]{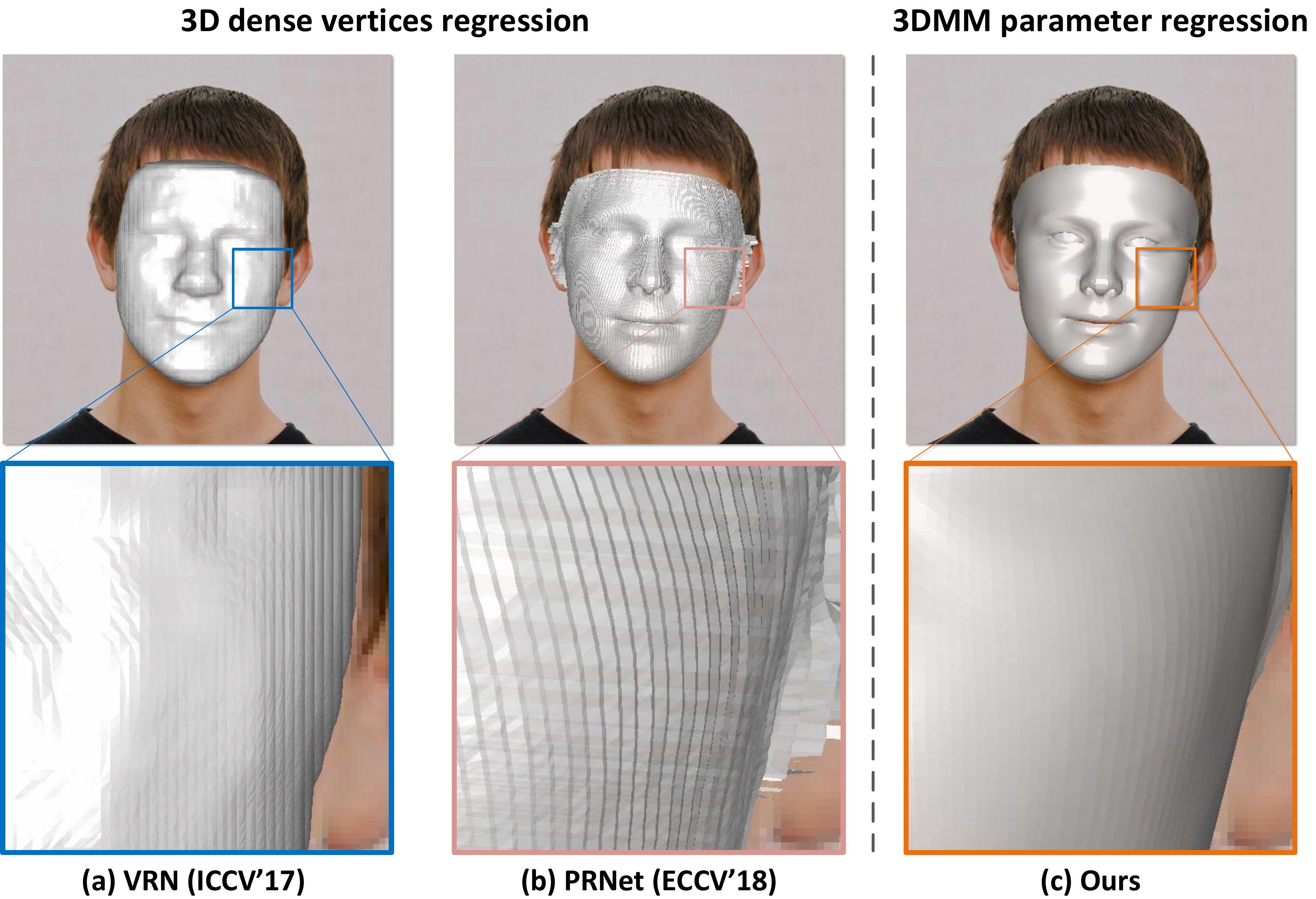}
  \caption{A result of PRNet~\cite{feng2018joint}, VRN~\cite{jackson2017large} and our 3DDFA-V2. The upper row is the dense mesh overlapped with the original image, the bottom row is the local details enlarged (better view in the electronic version). Local details show that the output mesh of PRNet is jagged and has checkerboard artifacts, VRN also has slight checkerboard artifacts, and our result is the smoothest.}
  \label{fig_jagged}
\end{figure}

The checkerboard artifacts of dense vertices regression~\cite{feng2018joint,jackson2017large} are shown in Fig.~\ref{fig_jagged}.

\section*{B. Impact of Dimension Reduction}
\begin{figure}[!h]
  \centering

  \includegraphics[width=0.625\textwidth]{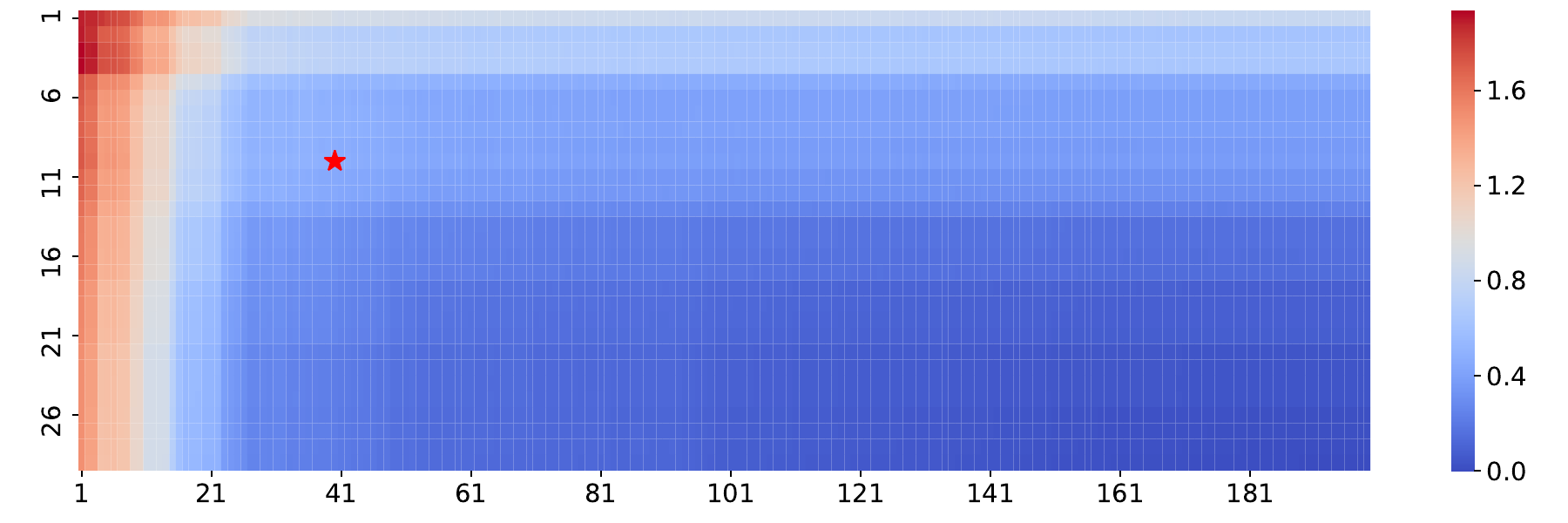}

  \caption{The 29$\times$199 heatmap of NME (\%) with different dimensions of shape and expression parameter (x-axis is shape, y-axis is expression). When the dimensions are set to [40, 10] (shown as the red star marker), the NME increase is about $0.4\%$, which is acceptable.}
  \label{fig_shape_exp}
\end{figure}

The NME error heatmap caused by different size of shape and expression dimensions is shown in Fig.~\ref{fig_shape_exp}.

\section*{C. Implementation Details}
Our experiments are based on PyTorch~\cite{paszke2017automatic}. During training, all faces are cropped and resized to 120$\times$120, then normalized by subtracting 127.5 and being divided by 128. We use SGD with a batch size $B$ of 128 to optimize the network, with the weight decay of 0.0005 and momentum of 0.9. For our model \textit{3DDFA-V2 (M+R+S)}, $k$ is 100 for the meta-joint optimization, and for the short-video-synthesis, each still image is synthesized with $n=8$ frames and the perturbation settings are: $\Delta s \in [0.95, 1.05]$, $\Delta \theta \in [-3^\circ, 3^\circ]$, $\Delta t1, \Delta t2 \in [-5, 5]$ pixels, $\Delta \phi, \Delta \gamma  \in[-5^\circ, 5^\circ]$.

\section*{D. Generalization and Scaling-up Ability}
\begin{figure}[!h]
  \centering
   \includegraphics[width=0.95\textwidth]{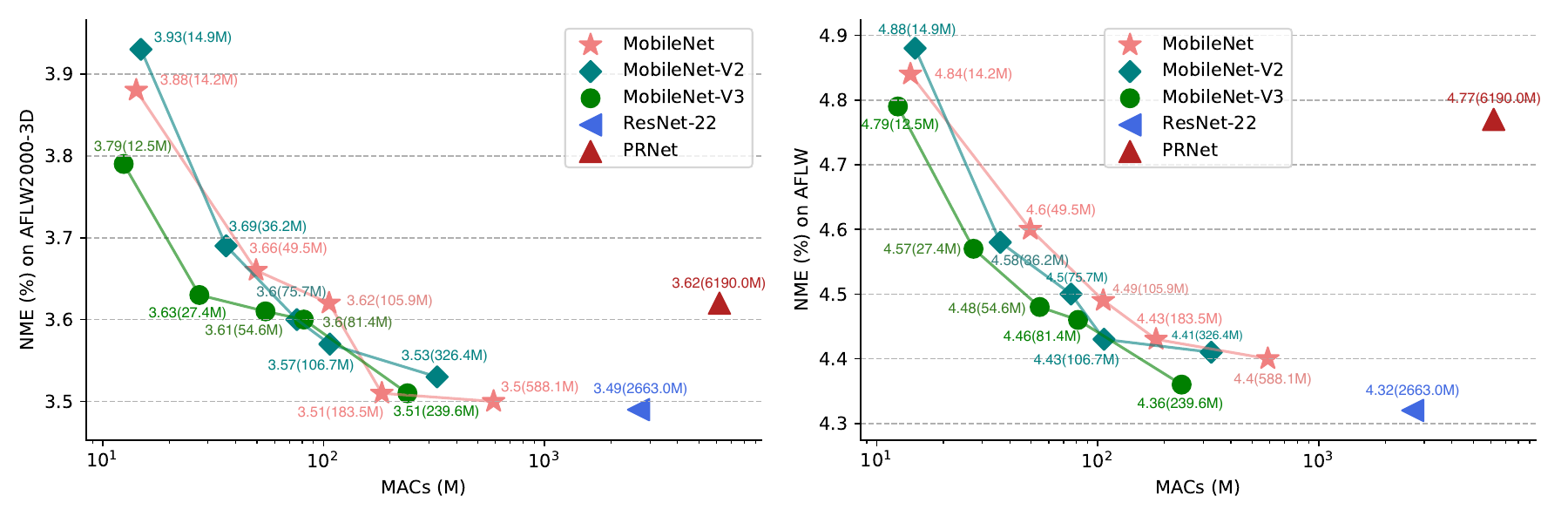}
  \caption{The trade-off between the computation complexity MACs and NME (\%) on AFLW2000-3D and AFLW. MobileNet, MobileNet-V2 and MobileNet-V3 (large mode) use multipliers 0.25, 0.5, 0.75 and 1 with input size 120 or 128 and the multiplier 1 with input size 224. ResNet uses 120. PRNet is shown here for comparison. Lower NME (\%) is better.}
  \label{fig_tradeoff}
\end{figure}

We compare the performance and speed with different architectures and scaling-up options in Table~\ref{tab_backbone} and Fig.~\ref{fig_tradeoff}. Note that the proposed methods are all applied on them.
The results in Table~\ref{tab_backbone} and Fig.~\ref{fig_tradeoff} reveal the generalization and scaling-up ability of our proposed methods: (i) when equipped with a more powerful backbone like ResNet-22, our methods perform better, which demonstrates the generalization ability across architectures; (ii) with different multipliers and input size, our methods show the great scaling-up ability. Users can choose the proper scaling-up option according to their need. Besides, MobileNet-V3~\cite{howard2019searching} performs better than MobileNet and MobileNet-V2~\cite{sandler2018mobilenetv2}, and MobileNet-V3 $\times$0.5 gives similar performance to PRNet with only 27.4M MACs, indicating that it is 225x faster than PRNet (6190M MACs) theoretically.

\begin{table}[!h]
\small
\centering
\caption{Comparisons of performance and speed on AFLW2000-3D, AFLW and Menpo-3D with different channel numbers and backbones. We ignore the reconstruction time (1ms in CPU) of 3D dense vertices in this table.}
\resizebox{0.95\textwidth}{!} {
\begin{tabular}{c||c|c|c||c|c|c}
\hline
\textbf{Backbone} & \textbf{AFLW2000-3D} & \textbf{AFLW}  & \textbf{Menpo-3D} & \textbf{Params} & \textbf{MACs} & \textbf{Inference Time (CPU)} \\ \hline
PRNet~\cite{feng2018joint} & 3.62 & 4.77 & 1.90 / 0.54 & 13.4M & 6190M & 175ms\\
PRNet $\times$0.25 & 4.77 & 6.54 & - & 0.84M & 434M & 48.7ms\\
PRNet $\times$0.125 & 5.24 & 7.06 & - & 0.21M & 134M & 38.4ms\\ \hline
ResNet-22 & \textbf{3.49} & \textbf{4.32} & \textbf{1.67} / \textbf{0.45} & 18.45M & 2663M & 67.5ms\\ 
MobileNet & \textbf{3.51} & \textbf{4.43} & \textbf{1.71} / \textbf{0.48} & 3.27M & 183.5M & 6.2ms \\ 
MobileNet $\times$0.75 & 3.62 & 4.49 & 1.74 / 0.50 & 1.86M & 105.9M & 4.2ms\\ 
MobileNet-V3 $\times$0.5 & 3.61 & 4.48 & 1.80 / 0.51 & 1.65M & 27.4M & 3.4ms\\ 
\hline
\end{tabular}
}
\label{tab_backbone}
\end{table}

\section*{E. Qualitative Results}
\begin{figure*}[!h]
  \centering
  \includegraphics[width=0.975\textwidth]{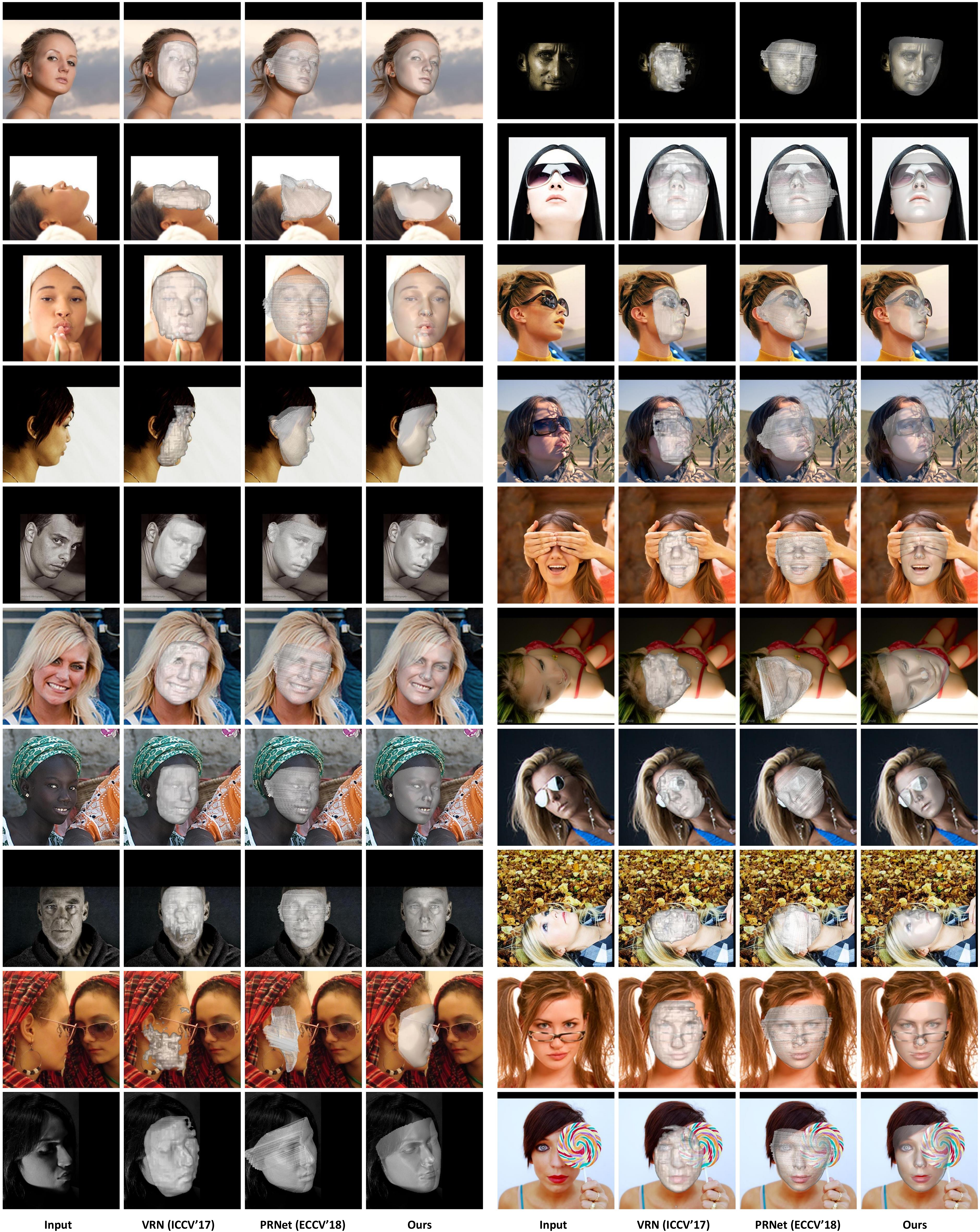}
  \caption{Qualitative results on AFLW2000-3D and AFLW. Our results are from the \textit{MobileNet (M+R+S)} model, which runs at over 50fps on a single CPU core. Please zoom in to see local details. (better view in the electronic version)}
  \label{fig_still}
\end{figure*}

We present more qualitative results (Fig.~\ref{fig_still}) for comparisons with VRN~\cite{jackson2017large} and PRNet~\cite{feng2018joint} on AFLW2000-3D and AFLW. The supplementary video presents 3D sparse and dense face alignment results.

\section*{F. Acceleration with ONNX Runtime\footnote{\url{https://github.com/microsoft/onnxruntime}}}
With the ONNX runtime and a single image as input, the inference speed of 3DDFA-V2\footnote{\url{https://github.com/cleardusk/3DDFA_V2}} is further accelerated to \textit{4.4ms} on a single CPU core or \textit{1.35ms} on four CPU cores (i5-8259U CPU @ 2.30GHz).

\end{document}